# REAL TIME CLUSTERING OF TIME SERIES USING TRIANGULAR POTENTIALS

Aldo Pacchiano[1]  and Oliver J. Williams[2]

[1]Massachusetts Institute of Technology
`aldopacchiano@gmail.com`
[2]Markham Rae LLP, London, UK. (The views and opinions expressed herein are those of the author and do not necessarily reflect the views of Markham Rae LLP.)
`oliver.williams@markhamrae.com`

## ABSTRACT

*Motivated by the problem of computing investment portfolio weightings we investigate various methods of clustering as alternatives to traditional mean-variance approaches. Such methods can have significant benefits from a practical point of view since they remove the need to invert a sample covariance matrix, which can suffer from estimation error and will almost certainly be non-stationary. The general idea is to find groups of assets which share similar return characteristics over time and treat each group as a single composite asset. We then apply inverse volatility weightings to these new composite assets. In the course of our investigation we devise a method of clustering based on triangular potentials and we present associated theoretical results as well as various examples based on synthetic data.*

## KEYWORDS

*Clustering, Expected Utility, Graphical Models, k-Clique Problem*

## 1. INTRODUCTION

A common problem in finance is the question of how best to construct a diversified portfolio of investments. This problem is ubiquitous in fund management, banking and insurance and has led to an extensive evolving literature, both theoretical and empirical. From an informal mathematical perspective the central challenge is to devise a method for determining weightings for a set of random variables such that *ex post* realisations of the weighted sum optimise some objective function *on average*. The objective function most typically used in financial economics is a concave *utility* function, hence from an *ex ante* pespective the portfolio construction problem is a matter of optimising so-called *expected utility*. Koller and Friedman provide a detailed discussion of utility functions and decision theory in the general machine learning context [1].

The theoretical literature analyses many alternative weighting strategies which can be distinguished based on such criteria as: (a) the investor's time horizon (e.g. does utility depend on realisations on a single time horizon in a 'one-shot' scenario or does uncertainty resolve over multiple time periods, affording the investor opportunities to alter portfolio composition dynamically?), (b) the nature of the information available to investors regarding the distribution of future returns (this may be extremely limited or highly-structured for mathematical expediency), and (c) the investor's particular utility function (where, for instance, it can be shown that curvature can be interpreted as representing the investor's risk-preferences [2]).

One of the most prominent theoretical results is the concept of *mean-variance efficiency* which has its roots in the work of Markowitz [3]: the idea is that in a one period model (under certain restrictive assumptions) if investors seek to maximize return and minimise portfolio variance, the optimal *ex ante* weighting vector $w$ is given by

$$w = \frac{1}{\lambda} \Omega^{-1} (\mu - r\iota) \tag{1}$$

where $\Omega$ is the covariance matrix of future returns, $\mu$ is the mean vector of expected returns, $\lambda$ is a risk-aversion parameter and $r$ is the risk-free rate of return [4]. A key aspect of this formula is the dependency on the inverse of the covariance matrix which is never known with certainty and will in practice be a forecast in its own right (and the same will be true for $\mu$ and quite possibly $r$). When deploying this formula in real-world investment, practitioners are divided over how to account for parameter uncertainty, with a number of alternative approaches in common usage (including ignoring uncertainty entirely).

Unfortunately it is widely recognised that the exact weightings in (1) have a sensitivity to covariance assumptions which is unacceptably high; in other words small changes in covariance assumptions can lead to large changes in prescribed weightings. Further significant concerns are raised by the fact that long time series are required to generate acceptable estimates for a large covariance matrix but financial returns series are notoriously non-stationary – it is therefore easy for an analyst to fall into the trap of thinking that they are applying prudent statistical methods when in reality their input data may be stale or entirely inappropriate. The forecasting of expected returns is also regarded as an exceptionally difficult task.

In these circumstances one strand of literature considers simpler weighting schemes which are predicated on relatively few assumptions; one prominent example, popular with practitioners, is the self-explanatory *equally-weighted* (or $\frac{1}{n}$) approach [5]. This method requires no explicit forecasts of correlation or returns and it can be shown that this is equivalent to mean-variance methods if the correlation between all possible pairs of investments is equal, along with all means and variances. Although this may be far from the truth it may be more innocuous to assume this than to suffer potentially negative effects of erroneous statistical forecasts and there is a body of empirical literature which demonstrates the efficiency of the approach [6]. Refinements to the basic method can include weighting each asset by the inverse of the forecast standard deviation of its returns (known as *volatility*) which allows some heterogeneity to be incorporated.

Nevertheless it is intuitively obvious that such a simple method presents potential dangers of its own, and is particularly inappropriate when the universe of alternative investments contains subgroups of two or more investments which are highly correlated with each other. Suppose, for instance, a portfolio of investments in world stock market indices which includes several alternative indices for the United States (e.g. Dow Jones, S&P 500, Russell 2000) but only single indices for other markets (e.g. the CAC-40 for France, FTSE-100 for UK, etc.). In this setting the $\frac{1}{n}$ approach may (arguably) significantly overweight US equities in comparison to each foreign market and in general regional weightings will be more dependent on the cardinality of available indices than any economic properties of the markets. In a systematic investment process it is clearly impractical to have analysts manually sift through investments to ensure an appropriate 'balance' (which defeats the object of a weighting algorithm) and indeed potential diversification benefits argue in favour of including a broad range of investments anyway.

The contribution of this paper is to explore potential weighting methods based on clustering,

such that highly 'similar' investments can be identified, grouped together and treated (for weighting purposes) as if they are a single 'composite' investment. By contrast, investments which exhibit relatively little similarity to each other are treated individually in their own right. Our focus here is on a process for identifying clusters rather than evaluation of *ex post* investment performance, which we leave for a separate analysis, and in fact we draw attention to the applicability of our methods to fields beyond finance where clustering may be required, e.g. well-known problems in biology, medicine and computer science. We also present an intriguing theoretical result arising from our work, which emphasises limitations of certain clustering techniques and may help to guide other researchers in their search for suitable methods.

The paper is organised as follows: in Section 2 we formally specify the problem at hand, in Section 3 we demonstrate spectral clustering as a preliminary benchmark approach and in Section 4 we explore an alternative method based on a graphical model where we propose a specific estimation technique involving *triangular potentials* and provide illustrative examples. Section 5 briefly considers extension to a more dynamic setting (via a Hidden Markov Model) and Section 6 concludes.

## 2. PROBLEM SPECIFICATION

**Definition 1** *Let* $n \in \mathbb{N}$, *define* $[n] = \{1, \cdots, n\}$ *the set of natural numbers from* $1$ *to* $n$.

Let $\{t_1\}, \cdots, \{t_n\}$ be $n$ time series, where $t_i = \{t_{i_1}, \cdots, t_{i_m}\}$ for $m, n \in \mathbb{N}$.

**Definition 2** *Clustering*.

A clustering of $\{t_1\}, \cdots, \{t_n\}$ is an equivalence relation $\sim$ over $[n] = \{1, \cdots, n\}$ such that:

1. Reflexivity: If $i \sim j$ then $j \sim i$.
2. Transitivity: If $i \sim j$ and $j \sim k$ then $k \sim i$.

**Definition 3** *Time dependent clustering*.

We say $i \stackrel{k}{\sim} j$ if $i$ and $j$ are clustered at time $k$.

Our aim is to find a sequence $\{\stackrel{k}{\sim}\}_{k=1}^{m}$, i.e. we allow the nature of the clustering relation to evolve over time.

We denote the *distance* between series at time $k$ as $d^k(t_i, t_j)$ for all $i, j$ and the similarity at time $k$ defined as $s^k(t_i, t_j)$. The functions $d^k, s^k$ are specified by the user of the algorithm and may be chosen based on prior domain-specific knowledge, or perhaps by a more systematic process of searching across alternative specifications guided by out-of-sample performance.

**Definition 4** *Distance Matrix*.

Given a family of time-dependent distance functions $\{d^k(\cdot, \cdot)\}_{k=1}^{m}$, we define a family of distance matrices as $D_{i,j}^k = \{d^k(t_i, t_j)\}$.

**Definition 5** *Similarity Matrix*.

Given a family of time-dependent similarity functions $\{s^k(\cdot, \cdot)\}_{k=1}^{m}$, we define a family of similarity matrices as $S_{i,j}^k = \{s^k(t_i, t_j)\}$.

**Definition 6** *Similarization function.*

We say $z : \mathbb{R}_+ \to [0,1]$ is a similarization function if for any distance function $d : \mathbb{R}^n \times \mathbb{R} \to \mathbb{R}_+$, $z \circ d$ is a valid similarity function.

In what follows we restrict our attention to reflexive and non-negative distance and similarity functions and thus to symmetric similarity and distance matrices. We will also use the variable $n$ to represent the number of data points observed at each time step when the clustering algorithm will be applied.

## 3. SPECTRAL CLUSTERING

Here we introduce the Spectral Clustering algorithm, which is suitable for data where the cluster structure does not change over time. Later in the paper we will compare the performance of our proposed approach with this benchmark method.

**Definition 7** *The Laplacian matrix $L$ of a similarity matrix $S$ is defined as follows:*

$$L = I - \mathsf{D}^{-\frac{1}{2}} S \mathsf{D}^{-\frac{1}{2}}$$

where $\mathsf{D}_{ii} = \sum_j S_{ij}$.

The most basic spectral clustering algorithm for bipartition of data is the Shi Malik bipartition algorithm which we describe below.

### 3.1. Shi Malik algorithm

Given $n$ items and a similarity matrix $S \in \mathbb{R}^{n \times n}$, the Shi Malik algorithm bipartitions the data into two sets $(B_1, B_2)$ with $B_1 \cup B_2 = [n]$ and $B_1 \cap B_2 = \varnothing$ based on the eigenvector $v$ corresponding to the second smallest eigenvalue of the normalized Laplacian matrix $L$ of $S$.

**Algorithm 1** *The Shi Malik bipartition algorithm:*

1. Compute the Laplacian from a similarity matrix.
2. Compute the second smallest eigenvalue and its corresponding eigenvector $v$.
3. Compute the median $m$ of its corresponding eigenvector.
4. All points whose component in $v$ is greater than $m$ are allocated to $B_1$, the remaining points are allocated to $B_2$.

Unfortunately the Shi Malik algorithm is not a dynamic procedure, i.e. it is not intended to identify an underlying cluster structure which is time-varying. However various clustering approaches are available which specifically seek to address this and we outline one such approach next.

### 3.2. A generalized spectral clustering approach

The following algorithm is an extension of the Shi Malik algorithm that can handle two or more clusters. It can be found at [7]. Given $n$ items and a similarity matrix $S \in \mathbb{R}^{n \times n}$ the goal of Dynamic Spectral Clustering is to find a clustering $\sim$ of $[n]$.

**Algorithm 2**   *Dynamic Spectral Clustering*

1. Compute the Laplacian of the similarity matrix.
2. Compute the Laplacian's eigenvalues and eigenvectors
3. Let $c$ be a desired number of clusters.
4. Find the eigenvectors of the corresponding eigenvalues found on the previous step. Let the corresponding $n \times c$ matrix be called $V$.
5. Rotate $V$, by multiplying it with an appropriate rotation matrix $R$ so each of the corresponding rows of $Z = VR$ have (ideally) only one nonzero entry. In reality the resulting matrix we will use the largest (in absolute value) entry of the matrix. $R$ is a rotation matrix in $R^{c \times c}$.
6. The cluster to which point $i$ is assigned is $\arg\max_{j \in \{1,\cdots,c\}} |Z_{i,j}|$.

In order to find an appropriate rotation matrix $R$, there is a theorem that guarantees that any rotation matrix $R \in R^{c \times c}$ can be written as a product $G_1 \cdots G_k$ where $k = \dfrac{c(c-1)}{2}$ and each $G_i$ equals a Givens rotation matrix.

Givens rotation matrices $G(i, j, \theta)$ are parameterized as follows:

$$G(i,j,\theta) = \begin{bmatrix} 1 & \cdots & 0 & \cdots & 0 & \cdots & 0 \\ \vdots & \ddots & \vdots & & \vdots & & \vdots \\ 0 & \cdots & \cos\theta & \cdots & -\sin\theta & \cdots & 0 \\ \vdots & & \vdots & \ddots & \vdots & & \vdots \\ 0 & \cdots & \sin\theta & \cdots & \cos\theta & \cdots & 0 \\ \vdots & & \vdots & & \vdots & \ddots & \vdots \\ 0 & \cdots & 0 & \cdots & 0 & \cdots & 1 \end{bmatrix}$$

Hence for each $G_i$ there is an associated angle $\theta_i$ and we represent these $k$ angles by the vector $\Theta \in R^{c(c-1)/2}$. In order to find the optimal $\Theta$ for a given number of clusters $c$, we use gradient descent on the following objective function:

$$\min_{\Theta} J = \sum_{i=1}^{n} \sum_{j=1}^{c} \left(\frac{Z_{ij}}{M_i}\right)^2$$

subject to the constraint

$$Z_{n \times c} = V_{n \times c} R(\Theta)_{c \times c}.$$

Following [7] we set $M_i = \max_j |Z_{ij}|$.

As suggested by [7], the optimal number of clusters can be obtained by choosing the value of $c$ that maximizes a scoring function given by

$$q(c,n) = 1 - \left(\frac{J}{n} - 1\right).$$

### 3.2.1. A dynamic clustering algorithm

Given a family of time dependent similarity functions $\{s^k(\cdot,\cdot)\}_{k=1}^m$ defining a family of similarity matrices $S_{i,j}^k = \{s^k(t_i, t_j)\}$, an optimal time-varying clustering structure $\overset{k}{\sim}$ can be estimated by applying Algorithm 2 at time $k$ using input similarity matrix $S_{i,j}^k$. Hence for time series data we propose the following algorithm:

**Algorithm 3** *Let* $\{t_1\}, \cdots, \{t_n\}$ *be* $n$ *time series. Where* $t_i = \{t_{i_1}, \cdots, t_{i_m}\}$ *for* $m, n \in \mathbb{N}$. *Let* $w \in \mathbb{N}$ *be a window parameter,* $d(\cdot,\cdot) : \mathbb{R}^w \times \mathbb{R}^w \to \mathbb{R}_+$ *be a distance function and* $z : \mathbb{R}_+ \to [0,1]$ *be a similarization function.*

1. Let $D_{ij}^m$ be the distance matrix having

$$D_{i,j}^m = d([t_{i_{m-w+1}}, \cdots, t_{i_m}], [t_{j_{m-w+1}}, \cdots, t_{j_m}])$$

for every pair $i, j \in [n]$.

2. Let $S_{i,j}^m$ be the similarity matrix having $S_{i,j}^m = z(D_{i,j}^m)$ for every pair $i, j \in [n]$.

3. Let $\overset{m}{\sim}$ be the clustering resulting from running Algorithm 2 with input similarity matrix $S_{i,j}^m$.

4. Output clustering $\overset{m}{\sim}$.

Extensions of this approach include considering a geometric decay factor in the distance computation, alternative distance functions and different similarization functions. We tried various combinations as shown in Table 1 but found no significant improvement on the stability of the resulting clusters.

We did not consider a scenario where the distance or similarization functions change through time although there may be certain applications where this might be appropriate.

Table 1: Alternative distance and similarity functions. The second similarity function is a generalization of the first.

| **Distances** | $L^1$ norm | $L^2$ norm |
|---|---|---|
| **Similarities** | $\exp\left(-\left\|\frac{x_1}{\|x_1\|} - \frac{x_2}{\|x_2\|}\right\|\right)$ | $\exp\left(-c \cdot \left\|\frac{x_1}{\|x_1\|} - \frac{x_2}{\|x_2\|}\right\|\right)$ set to zero when it achieves values less than $\lambda$ for different combinations of |

|  |  | $c, \lambda$ |
|---|---|---|

## 3.3. Overview

We present the performance of this algorithm in Figure 1. Some of the observed characteristics of this method are the following:

• The resulting clustering values are notably sensitive to the similarity function used in the model.

• The clustering structure estimated by this method tends to be relatively unstable over time. Although in some applications this may be plausible, in the context of financial time series we have a strong prior belief that clusters typically arise due to common factors relating to economic fundamentals (e.g. similar commodities, currency pairs belonging to close trading partners, etc.) which would tend to change very slowly relative to the frequency of market data.

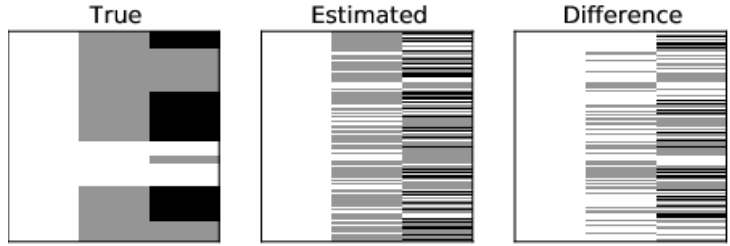

Figure 1: Performance of the spectral clustering algorithm on 5,000 periods of synthetic data with $n = 3$: at each time step we generate random standard normal variates which are common to each cluster, then for each of the 3 returns we add independent Gaussian noise with a relatively small variance. The members of each cluster therefore have a large portion of randomness in common, but each observation also includes its own independent noise. The cluster structure is randomly changed over time and represented by coloured bars in each row, i.e. all columns with the same colour belong in the same cluster.

## 4. GRAPHICAL MODEL APPROACH

Instead of representing clusterings as a binary matrix $Z(i,j)$ such that $Z(i,j) = 1$ if $i \in$ cluster $j$ as the authors of [8] do, we approach the problem in a different way. Consider a symmetric ($C_{i,j} = C_{j,i}$) family of Bernoulli random variables, $\mathbf{C} = \{C_{i,j}\}_{i,j \in [n]}$ such that:

$$C_{i,j} = 1 \text{ if } i,j \text{ are in the same cluster, or 0 otherwise.}$$

We wish to learn a distribution over the ensemble $\mathbf{C} = \{C_{i,j}\}$. The model we will use in this paper is the following:

$$\mathbf{C} \to S$$

where $S$ is a similarity matrix; in other words, we consider that the observed similarity between a pair of points will come from one of two distributions, depending on whether or not the two points belong to the same cluster.

In what follows it will be useful to think of the matrix $\{C_{i,j}\}$ as an adjacency matrix. The resulting graph $G = (V, E)$ where $V = [n]$ and $E = \{(i, j) \mid C_{i,j} = 1\}$, has an edge between every two nodes that are in the same cluster. Learning a distribution over $\{C_{i,j}\}$ can be thought of as learning a distribution over the set of undirected graphs $(V, E)$ with $V = [n]$.

The goal of this section is to compute the following posterior:
$$P(\mathbf{C} \mid S).$$

The algorithms we present here output $\arg\max_{\mathbf{C} \in \mathcal{C}} P(\mathbf{C} \mid S)$, the MAP estimator for the posterior.

A short algebraic manipulation (Bayes Theorem) yields:
$$P(\mathbf{C} \mid S) = \frac{P(S \mid \mathbf{C}) \cdot P(\mathbf{C})}{P(S)}.$$

Since $S$ is fixed:
$$\arg\max_{\mathbf{C} \in \mathcal{C}} P(\mathbf{C} \mid S) = \arg\max_{\mathbf{C} \in \mathcal{C}} P(S \mid \mathbf{C}) P(\mathbf{C}).$$

In the following two sections we present different models for inference on the ensemble $\mathbf{C}$, their performance and their relationship to clusterings.

The training data will be:

1. A set of similarity matrices $\{S_{i,j}^k\}_{i=1}^m$.

2. The set of corresponding clusterings $\{\tilde{}\}^k$ produced via a clustering algorithm such as the ones described earlier in Section 3.

### 4.1. Exponential model

As a starting point we propose the following model for the ensemble $\mathbf{C}$, in which we impose conditional independence assumptions between observed similarities. We therefore assume the following factorization:

$$P(\mathbf{C} \mid S) = \frac{1}{Z(S)} \Phi(\mathbf{C}, S) = \frac{1}{Z(S)} \prod \Psi_{i,j}^1(C_{i,j}, S_{i,j}) \Psi^2(C_{i,j})$$

$$Z(S) = \sum_{\mathbf{C} \in \mathcal{C}} \prod \Psi_{i,j}^1(C_{i,j}, S_{i,j}) \Psi^2(C_{i,j})$$

In this model we assume $\Psi_{i,j}^1(C_{i,j}, S_{i,j}) = P(S_{i,j} \mid C_{i,j})$, and $\Psi^2(C_{i,j}) = P(C_{i,j})$. This is equivalent to assuming full pairwise independence of the variables $C_{i,j}$ and the conditionals $P(S_{i,j} \mid C_{i,j})$.

For implementational purposes we assume $C_{i,j} \to S_{i,j}$ are exponentially distributed and the $C_{i,j}$ are Bernoulli random variables.

### 4.1.1. Training

$\{\overset{k}{\sim}\}$ can be translated into a training sequence of ensemble values $\{C^k\}$ via the transformation $C^k_{i,j} = 1$ if $i \overset{k}{\sim} j$. Because of the independence assumptions underlying this model, the ML estimate for the posterior distribution of the ensemble can be computed by obtaining the ML estimate for each of the distributions $P(S_{i,j} | C_{i,j})$ and $P(C_{i,j})$. The ML estimate for the rate parameter of $P(S_{i,j} | C_{i,j})$ equals the inverse of the sample mean, and the ML estimate for the mean of $P(C_{i,j})$ equals the sample frequency of $C_{i,j} = 1$. More formally:

**Observation 1** *Define* $\lambda_{i,j} = \frac{1}{m} \sum_k S^k_{i,j}$. *And let* $p_{i,j} = \frac{1}{m} \sum_k C^k_{i,j}$.

The ML estimator of the parameters for the posterior distribution $P(\mathbf{C} | S_{i,j}) = \frac{1}{P(S)} P(S_{i,j} | \mathbf{C}) P(\mathbf{C})$ has $P(S_{i,j} | C_{i,j}) \sim \exp(\lambda_{i,j})$ and $P(C_{i,j} = 1) = p_{i,j}$.

### 4.1.2. Prediction

Prediction under this model is performed by finding the MAP assignment for the ensemble $C^*$ and turning it into a clustering. $C^*$ is obtained by maximizing each likelihood $P(S_{i,j} | C_{i,j}) P(C_{i,j})$ independently:

$$C^*_{i,j} = \arg \max_{C_{i,j} \in \{0,1\}} P(S_{i,j} | C_{i,j}) P(C_{i,j}).$$

For the ensemble assignment $C^*$ we output a clustering composed of a cluster for each connected component of the graph corresponding to $C^*$. Results are presented in Figure 2.

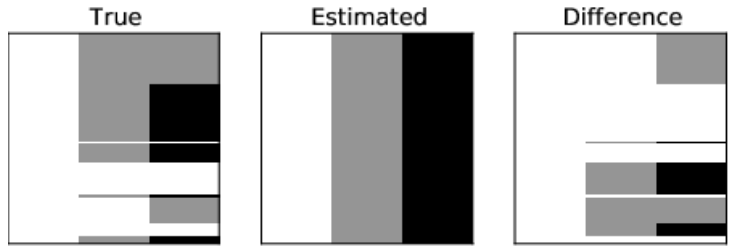

Figure 2: Performance of the exponential model on 5,000 periods of synthetic data with $n = 3$.

The prediction algorithm is linear.

### 4.1.3. Limitations

Consider the following joint posterior distribution over clusterings of $\{1,2,3\}$.

$$p(\cdot) = \begin{cases} 0.1 & \text{if } \cdot = (1,2,3) \\ 0.41 & \text{if } \cdot = (1,2),(3) \\ 0.41 & \text{if } \cdot = (1,3),(2) \\ 0 & \text{if } \cdot = (2,3),(1) \\ 0.17 & \text{if } \cdot = (1),(2),(3) \end{cases}$$

The marginals $p((1,2)), p((2,3)) > 0.5$. The current algorithm will output $(1,2,3)$.

### 4.2. Triangular Potentials

The main limitation of the approach described in the previous section is that there is potential for spurious large clusters to emerge solely from the independent optimization of the potentials. If the marginal probability $p_{i,j}$ is large, it is likely that the MAP of the ensemble $C$ will have $C_{i,j} = 1$ regardless of the values of any of the other similarities $S_{k,m}$ or clustering assignments $C_{k,m}$. It is also possible for the algorithm to suggest cluster shapes which are intuitively implausible (and do not conform to prior notions of cluster structure which may be appropriate to a particular domain); we illustrate this in Figure 3.

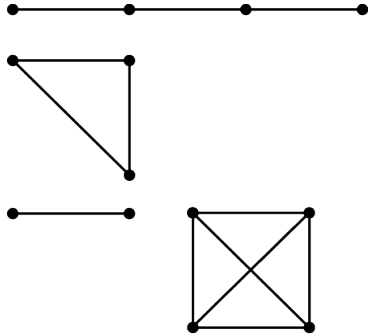

Figure 3: Alternative cluster structures: (top to bottom) the first implausible configuration is ruled-out by the use of *triangular potentials*, however the second and third configurations are possible (which is our deliberate intention).

We therefore proceed to address these issues by a modification to the basic model as described by the following observations:

**Observation 2** $C$ *is a valid clustering* $\sim$ *if, for all triplets of distinct numbers* $i, j, k \in [n]$, $C_{i,j} = C_{j,k} = 1 \Rightarrow C_{k,i} = 1$.

**Observation 3** $C$ *is a valid clustering if the graph whose adjacency matrix equals* $C$ *is*

*composed of a disjoint union of cliques.*

In this section we assume the following factorization:

$$P(\mathsf{C}|S) = \frac{1}{Z(S)} \Phi'(\mathsf{C}, S)$$

$$= \frac{1}{Z(S)} \left( \prod_{i,j} \Psi^1_{i,j}(C_{i,j}, S_{i,j}) \Psi^2_{i,j}(C_{i,j}) \right) \prod_{i,j,k} \Psi^3_{i,j,k}(C_{i,j}, C_{i,k}, C_{j,k}) \qquad (2)$$

$$Z(S) = \sum_{\mathsf{C} \in \mathcal{C}} \left( \prod_{i,j} \Psi^1_{i,j}(C_{i,j}, S_{i,j}) \Psi^2_{i,j}(C_{i,j}) \right) \prod_{i,j,k} \Psi^3_{i,j,k}(C_{i,j}, C_{i,k}, C_{j,k})$$

where

$$\Psi^3_{i,j,k}(C_{i,j}, C_{i,k}, C_{j,k}) = \begin{cases} 0 & \text{if } C_{i,j} = C_{i,k} = 1, C_{j,k} = 0 \\ 0 & \text{if } C_{i,j} = C_{j,k} = 1, C_{i,k} = 0 \\ 0 & \text{if } C_{i,k} = C_{j,k} = 1, C_{i,j} = 0 \\ 1 & \text{otherwise} \end{cases}$$

This has the effect of turning $\Phi(\mathsf{C}, S)$ into a potential function $\Phi'(\mathsf{C}, S)$ such that all the assignments of the joint distribution of the ensemble $\mathsf{C}$ with a nonzero probability are valid clusterings.

### 4.2.1. Training algorithm

We use the same construction for the univariate and bivariate potentials as the one used in the previous section. The distribution over clusterings will vary because the triangular potentials restrict the mass of the distribution to the space of valid clusterings. It is of course also possible to add other potentials relating different sets of clustering variables although we leave that direction for future research.

### 4.2.2. Prediction algorithms

This model can be thought of as an undirected graphical model with variables $\{C_{i,j}\}$ for $i < j$ and $i, j \in [n]$ and edges $C_{i,j}, C_{i,k}$, $C_{i,j} C_{j,k}$, and $C_{j,k}, C_{i,j}$ for all $i < j < k$. If the variable $C_{i,j}$ is identified with the point $(i, j)$, then there is an edge between every two variables on the same vertical line and between every two variables on the same horizontal line.

We tackle the problem of obtaining the MAP assignment over clusterings under this model using either the Elimination Algorithm or MCMC. To obtain an estimate for the MAP assignment using MCMC we sample from the posterior and output the clustering arrangement which appears most often. The MCMC chain construction is described in the next section.

By construction there is a clique of size $n-1$ along the horizontal line $(1, i)$ for $i = 2, \cdots, n$ As a consequence, the elimination algorithm has an exponential running time over this graphical model. Similarly, there are no easy theoretical guarantees for the performance of the MCMC method. In particular, it is possible for the probability mass over the optimal assignment to be so small that there are no concentration inequalities to guarantee that the proposed algorithm will output the MAP with high probability in polynomial time.

In the following section we show this behavior is not only a result of the graphical model formulation or our proposed algorithm but an intrinsic limitation of the model itself.

### 4.3. Results and Limitations

We next apply the classic sumproduct algorithm or the MAP elimination algorithm to find the best clustering, with results shown in Figure 4, however the drawbacks are that this solution becomes intractable as the number of products becomes large. The elimination algorithm could be worst case $2^{n^2}$ which becomes intractable quite fast.

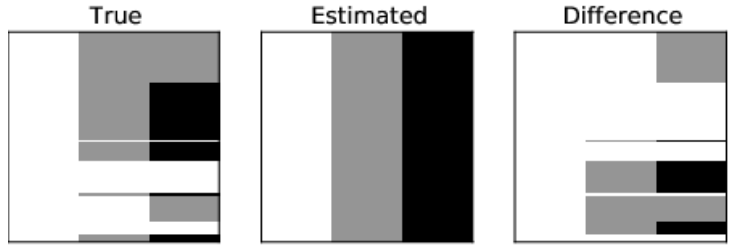

Figure 4: Performance of the graphical model with triangular potentials; $n = 3$.

#### 4.3.1. Theoretical limitations

Let $\hat{p}_{i,j} \in [0,1]$ be an ensemble of probabilities with $i \neq j$ such that $\hat{p}_{i,j} = \hat{p}_{j,i}$ and $i, j \in [n]$. Define a distribution over simple graphs via

$$P(G) = \left(\prod_{(i,j) \in E} \hat{p}_{i,j}\right)\left(\prod_{(i,j) \notin E} (1 - \hat{p}_{i,j})\right) \forall G = (V,E), V = [n].$$

Let $\hat{P}(G) = P(G \mid G \text{ is a disjoint union of cliques})$.

It is easy to see that finding the MAP assignment for the distribution defined via Equation (2) is equivalent to finding the MAP assignment for $\hat{P}(G)$ with:

$$\hat{p}_{i,j} = \frac{P(S_{i,j} \mid C_{i,j} = 1)P(C_{i,j} = 1)}{P(S_{i,j} \mid C_{i,j} = 1)P(C_{i,j} = 1) + P(S_{i,j} \mid C_{i,j} = 0)P(C_{i,j} = 0)}$$

Since it is conceivable that any arrangement of the values $\hat{p}_{i,j}$ can result from the training data, the two problems are equivalent.

In what follows we talk interchangeably of the MAP assignment $\{\hat{q}^*_{i,j}\}$ of $\hat{P}(G)$ $\left(q^*_{i,j} \in \{\hat{p}_{i,j}, 1 - \hat{p}_{i,j}\}\right)$ and the graph $G^* = (V^*, E^*)$ defined by $V^* = [n]$ and $E^* = \{\{i,j\} \mid \hat{q}^*_{i,j} = \hat{p}_{i,j}\}$. The complement of $G^*$ contains all those pairs $\{i,j\}$ for which

$$\hat{q}_{i,j}^* = 1 - \hat{p}_{i,j}.$$

**Theorem 1** *If there is a polynomial time algorithm for finding the MAP assignment over* $\hat{P}(G)$ *then P = NP.*

*Proof.* Let $\mathsf{A}(\{\hat{p}_{i,j}\})$ be an algorithm for finding the MAP over the distribution $\hat{P}(G)$ as defined by $\{\hat{p}_{i,j}\}$. We show $\mathsf{A}$ can be used to construct an algorithm for solving the $k-$clique problem. The $k-$clique problem is the problem of deciding whether a graph $G = (V, E)$ has a clique of size $k$ where both $G$ and $k$ are inputs to be specified. If $\mathsf{A}$ was polynomial, the algorithm we propose for $k-$clique would run in polynomial time. Because $k-$clique is NP complete we conclude the existence of $\mathsf{A}$ would imply P = NP. The following algorithm solves $k-$clique:

**Algorithm 4** *Inputs:* $<G, k>$.

Let $N > \max\{2 \cdot |E| + 2, 2|V|\}$ and $q > \dfrac{1}{2}$.

1. Construct $G' = (V', E')$ with $V' = V \cup V_N$ and $E' = E \cup \{\{v_1, v_2\} \mid v_1 \in V, v_2 \in V_N\} \cup \{\{v_1, v_2\} \mid v_1, v_2 \in V_N, v_1 \neq v_2\}$. The edges of $G'$ equal all edges in $G$, plus all possible edges between $V$ and $V_N$ and all possible edges among elements of $V_N$.

2. For all pairs $i, j \in [|V'|]$ define:
$$\hat{p}_{i,j} = \begin{cases} 0 & \text{if } i, j \notin E \\ q & \text{otherwise} \end{cases}$$

3. Let $\hat{E}^*$ be the output edges in the MAP assignment from $\mathsf{A}(\{\hat{p}_{i,j}\})$.

4. If $|\text{MaxClique}(\hat{E}^* \cap E)| \geq k$ output $1$, else output $0$. This step runs in polynomial time because every connected component of $\hat{E}^* \cap E$ is a clique graph.

The probability of the MAP assignment equals

$$\left(\prod_{(i,j) \in \hat{E}^*} \hat{p}_{i,j}\right)\left(\prod_{(i,j) \notin \hat{E}^*} (1 - \hat{p}_{i,j})\right)$$

which can be written as a product $P_N^* \cdot P_{N \times V}^* P_V^*$ of the product of the chosen probabilities of pairs belonging to $V_N \times V_N$, a cross component of probabilities from $V_N \times V$ and a component of probabilities from $V \times V$. By construction, the edges in $V \times V$ but not in $E$ are not chosen. The MAP restricted to $V_N$ and $V$ is a disjoint union of cliques. Because

$P_V^* \leq q^{|E|} < q^{N/2-1}$ we can conclude:

1. The MAP assignment restricted to $V_N$ must be a complete graph: Suppose the MAP restricted to $V_N$ had more than one component, say $K_1, \cdots, K_r$, $|K_1| \geq \cdots \geq |K_r|$ with $K_1^1, \cdots, K_r^1$ their (possibly empty) corresponding clique intersections in $V$. It can be shown via the rearrangement inequality that the MAP must have $|K_1^1| \geq \cdots \geq |K_r^1|$. Let MAP1 be the assignment obtained via joining $K_1, \cdots, K_r$ into $K_N$ (the complete graph on $V_N$) and reconnecting all to $K_1^1$. If $r \geq 2$, a simple counting argument shows that $\text{edges}(K_N) - \sum_{i=1}^{r} \text{edges}(K_i) \geq \frac{N}{2}$. The latter, and $|K_1^1| \geq \cdots \geq |K_r^1|$ imply that $P(\text{MAP1}) > P(\text{MAP})$, a contradiction.

2. The $V_N \times V$ edges must connect $V_N$ with one of the largest cliques of $G$.

The correctness of the algorithm follows. The algorithm above runs in polynomial time, provided $A$ is in P.

## 5. EXTENSIONS

### 5.1. HMM

Because the training procedure we propose is done over fully annotated data, more sophisticated and time-dependent models can be explored. We propose a generalization of the previous models via an HMM.

In this model, each hidden state is a clustering and the transition probabilities are obtained from the sampled frequencies of the transitions in the training phase. When the hidden states of the training data are known, the ML estimate of the transition probabilities of an HMM equals the transitions sample frequencies.

The results of applying this method are shown in Figure 5, where it is apparent that relatively good performance is achieved.

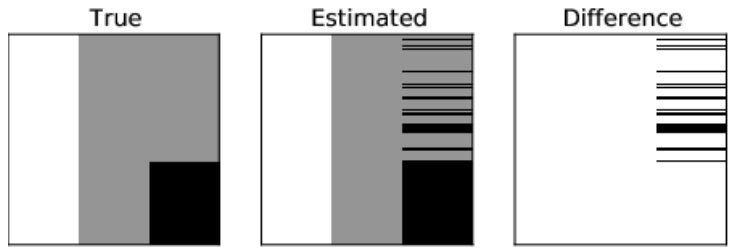

Figure 5: Performance of the HMM clustering algorithm on 2,000 periods of synthetic data with $n = 3$.

The version implemented here is hard-coded for only 3 series and therefore only 5 possible clustering states. The length of the chain can be adjusted as desired.

## 5.2. Coagulation Fragmentation

The underlying chain for the MCMC sampler uses a fragmentation coagulation process to walk over clusterings. At each step, the chain either selects a random cluster, and divides it into two, or selects two random clusters and joins them together. The acceptance/rejection probabilities can be computed with respect to any coagulation fragmentation process. In our implementation, we pick either a uniform random cluster and a random bipartition of it (fragmentation), or a uniform random pair of clusters (coagulation). We believe the mixing time of this process should be fast as it is related to a coagulation fragmentation process known as the random transposition walk. Diaconis and Shahshahani provided a polynomial upper bound for this walk's mixing time [9].

### 5.2.1. Alternative model

We believe a worthwhile alternative to the ideas described above is to represent the clustering evolution as an HMM on fragmentation-coagulation parameters: the simplest model having only two parameters $(p,q)$, one controlling the probability of fragmentation and the other controlling the probability of coagulation. If the number of fragmentation-coagulation parameters is small, inference could be tractable.

## 6. CONCLUSIONS

Our intention in this paper has been to show how various clustering methods can be applied to datasets which arise in financial markets. We have documented the process by which we analysed the problem and considered a method for determining clusters using *triangular potentials*. This latter method can be computationally intensive and we have provided some preliminary theoretical results concerning its limitations. However, notwithstanding these considerations, we have found promising empirical results from applying the method to simulated datasets and we look forward to extending this to real-world data in due course.

In future work we aim to extend the idea to a setting where we place a non-uniform prior on clusterings, e.g. if expert knowledge suggests that a group of investments are likely to share similar return characteristics then we can configure potentials such that appropriate weighted links are established among these products.

There is also considerable scope to investigate efficiency improvements to the MCMC estimation process, based on the particular structure of potentials in this context.

**Authors**

Aldo Pacchiano is a Bachelor of Science in Computer Science and Mathematics from MIT, with a masters degree in Mathematics from Cambridge University and a recently obtained Masters of Engineering degree from MIT; his interests span the areas of Machine Learning, Computational Biology and he has a special interest in quantitative finance.

Oliver Williams is a quantitative investment specialist with interests including financial economics, asset pricing and systematic trading; he has co-authored a number of papers in this area and his career has been spent in investment banking and asset management. He holds an MA in Computer Science and Management Studies, MPhil and PhD in Financial Economics from Cambridge University.